\RequirePackage{fix-cm}
\documentclass[twocolumn]{svjour3}          
\smartqed  
\usepackage{microtype}
\usepackage{graphicx}

\usepackage[caption=false]{subfig}
\usepackage{hyperref}
\usepackage{amsmath}
\usepackage{amsfonts}
\usepackage{amssymb}
\usepackage{enumerate}
\usepackage{color}
\usepackage[shortlabels]{enumitem}
\usepackage{todonotes}
\usepackage[normalem]{ulem} 

\definecolor{myblue}{RGB}{0, 50, 255}

\begin{document}

\title{A trained humanoid robot can perform human-like crossmodal social attention and conflict resolution}




\author{Di Fu$^{1,2,3}$          \and
        Fares Abawi$^{3}$  \and
        Hugo Carneiro$^{3}$ \and
        Matthias Kerzel$^{3}$ \and
        Ziwei Chen$^{1,2}$ \and
        Erik Strahl$^{3}$ \and
        Xun Liu$^{1,2}$ \and
        Stefan Wermter$^{3}$
}

\authorrunning{Di Fu et al.} 

\institute{
\\
$^{1}$D.~Fu, Z.~Chen, X.~Liu are with CAS Key Laboratory of Behavioural Science, Institute of Psychology, Beijing, China. {\tt\small czwscda2015@163.com}, {\tt\small liux@psych.ac.cn}\\
$^{2}$D.~Fu, Z.~Chen, X.~Liu are with the Department of Psychology, University of Chinese Academy of Sciences, Beijing, China.
\\
$^{3}$D.~Fu, F.~Abawi, H.~Carneiro, M.~Kerzel, E.~Strahl, S.~Wermter are with the Department of Informatics, University of Hamburg, Hamburg, Germany {\tt\small \{di.fu, fares.abawi, hugo.carneiro, matthias.kerzel, erik.strahl, stefan.wermter\}@uni-hamburg.de}
}


\date{Received: date / Accepted: date}

\maketitle

\begin{abstract}
{\color{black}To enhance human-robot social interaction, it is essential for robots to process multiple social cues in a complex real-world environment. However, incongruency of input information across modalities is inevitable and could be challenging for robots to process. To tackle this challenge, our study adopted the neurorobotic paradigm of crossmodal conflict resolution to make a robot express human-like social attention.} A behavioural experiment was conducted on 37 participants for the human study. We designed a round-table meeting scenario with three animated avatars to improve ecological validity. Each avatar wore a medical mask to obscure the facial cues of the nose, mouth, and jaw. The central avatar shifted its eye gaze while the peripheral avatars generated sound. Gaze direction and sound locations were either spatially congruent or incongruent. We observed that the central avatar's dynamic gaze could trigger crossmodal social attention responses. In particular, human performances are better under the congruent audio-visual condition than the incongruent condition. Our saliency prediction model was trained to detect social cues, predict audio-visual saliency, and attend selectively for the robot study. After mounting the trained model on the iCub, the robot was exposed to laboratory conditions similar to the human experiment. While the human performances were overall superior, our trained model demonstrated that it could replicate attention responses similar to humans.

\end{abstract}

\section{Introduction}
\label{sec:intro}
Robots are increasingly becoming an integral part of daily life. 
{\color{black}It is essential for robots to behave as social actors capable of processing multimodal social cues, enriching interactions with humans. Moreover, to understand humans' intentions, it is crucial to explore how they process information and the underlying cognitive mechanisms behind it~\cite{nocentini2019survey}. The need for such solutions encourages the design of socially functional robots to meet more significant challenges and difficulties in human-robot communication.} 

The current study adopts a dynamic variant of the gaze-triggered Posner cueing paradigm~\cite{posner1980attention} for testing the attentional orienting effect of eye gaze on auditory target detection. We construct a synthetic scenario using the framework introduced by Parisi et al.~\cite{parisi2018neurorobotic} and Fu et al.~\cite{fu2018assessing} to study crossmodal spatial attention for sound localisation. 
{\color{black}In the aforementioned studies, a 4-avatar round-table meeting scenario experiment was conducted on human participants.} During the task, lip and arm movements {\color{black}were} used as the visual cues, either spatially congruent or incongruent with the auditory target. Our previous findings indicated that lip movement {\color{black}was} more salient than arm movement, implying a stronger visual bias on auditory target localisation. This is due to the physical association between lip movement and speech~\cite{yeung2013lip}. Furthermore, previous research also revealed that head orientation {\color{black}was} a primary social cue for triggering the reflexive attention of an observer~\cite{langton2000eyes}. To align our experimental setup with the Posner gaze-cueing task, we reduce the number of avatars to three. The central avatar shifts its eyes with a slight tilt in its head and upper body posture towards the direction of gaze. To avoid distractions from lip movements, all three avatars wear medical masks to obscure their faces partially. The current social norm inspires this task design. In multiperson social contexts during the COVID-19 pandemic, the use of medical masks is common. Research shows that wearing masks decreases both adults' and children's face recognition abilities~\cite{stajduhar2022face,gori2021masking}. As a result, humans have to rely on gaze cues to compensate for the lack of lip movement in identifying social intentions~\cite{dalmaso2021face}.
    
For the robotic experiment in this work, an iCub head is used to emulate human social attention~\cite{nummenmaa2009}. We modify the Gated Attention for Saliency Prediction (GASP) model~\cite{abawi2021} and mount it on the iCub head to predict crossmodal saliency. GASP can detect multiple social cues, producing feature maps for each. These maps are prioritised based on a weighting mechanism to mitigate stronger cues. Following the weighting stage, the features are sequentially integrated, and the model is trained on eye tracking data to predict saliency. The iCub gaze movements are based on the saliency density maps predicted by the GASP model.
    
We define two goals for our current study. First, we aim to detect human responses for a crossmodal social attention task with dynamic stimuli to determine the eye gaze orienting effect on sound localisation. Second, we emulate human behavioural patterns using a humanoid robot, running a social attention model which is tested in similar laboratory conditions. Thus, human and robot responses are compared under congruent and incongruent audio-visual localisation conditions in the gaze-cueing task. {\color{black}In this study, the Stimulus-Response Compatibility (SRC) effect~\cite{proctor2006stimulus} is measured to detect the conflict resolution ability of the participants and the iCub robot. This effect occurs when stimulus and response in an SRC paradigm are spatially incongruent. Participants show poorer
performance (e.g., lower accuracy and slower response to stimuli) under incongruent conditions compared with congruent conditions~\cite{ambrosecchia2015spatial}. Larger SRC effects indicate weaker conflict resolution ability~\cite{liu2018neurodevelopment}.} {\color{black}Previous research also set a neutral condition as a baseline to distinguish whether irrelevant or incongruent stimuli cause the SRC effect entirely~\cite{schuller2004perception}. If there is no significant difference between participants' performance under neutral and congruent conditions, the SRC effect comes from irrelevant or incongruent stimuli interference. If the performance of the neutral condition is significantly worse than the congruent condition, it means that congruent stimuli have a facilitation effect on conflict processing~\cite{macleod1991half,kornblum1995stimulus}. Thus, we set a neutral condition to study whether there is an interference or facilitation effect where the central avatar does not shift its eyes, head, or upper body in any direction in the current study.}

According to our research goals, the current study proposes the following hypotheses:\\



{\color{black}In the human experiment:}

{\color{black}\textbf{H1:} Eye gaze can trigger the attentional orienting effect, which leads to better performances with the congruence of gaze direction and auditory targets.} 

{\color{black}\textbf{H2:} For the neutral condition, no irrelevant
visual stimulus shows up before the auditory target. We assume that participants' performance in the neutral condition might be intermediate between congruent and incongruent conditions. More specifically, no significant difference between performance under congruent and neutral conditions, suggests that the SRC effect is from the interference of the incongruent condition.} 

{\color{black}In the robot experiment:}

{\color{black}\textbf{H3:} Modelling the reflexive attentional orienting effect is achievable by integrating a binaurally aware auditory localiser for estimating the direction of sound arrival.}

{\color{black}\textbf{H4:} A neurocognitive model trained on human eye fixations can result in a robot attentional orientation consistent with human responses under the congruent, incongruent, and neutral conditions.}

{\color{black} To test the validity of these hypotheses, the article is structured into two parts. The first part focuses on how humans behave in a crossmodal conflict task triggered by eye gaze as a visual cue. Background on the use of eye gaze as a social cue is provided in Section~\ref{sec:background}. The full description of the experiment performed with human participants and the results achieved are provided in {\color{black}Section~\ref{sec:human_experiment}}. The following part of the article focuses on whether a robot can behave similarly to a human in the same experimental scenario. For that, a description of GASP, the attention mechanism used by the robot, is presented in Section~\ref{sec:gasp}, and the setup of the robotic experiment, as well as a comparison between the performances of the robot and those of the human participants, are presented in Section~\ref{sec:icub}. Finally, Section~\ref{sec:discussion} offers a discussion on the achieved results, and Section~\ref{sec:futurework} indicates potential future research directions.}

\section{Background and Related work}
\label{sec:background}
\subsection{Social Attention}
{\color{black}Social attention is the ability to follow others' eye gaze and infer where and what they are looking at~\cite{birmingham2009human}.} Social attention is the fundamental function of sharing and conveying information with other agents, contributing to the functional development of social cognition~\cite{mundy2007attention}. Social attention allows humans to quickly capture and analyse others' facial expressions, voices, gestures, and social behaviour, so that they can participate in social interaction and adapt within society~\cite{langton2000eyes,laube2011cortical}. Furthermore, this social function enables the recognition of others' intentions and the capture of relevant occurrences in the environment (e.g., frightening stimuli, novel stimuli, and reward)~\cite{nummenmaa2009}. The neural substrates underlying social attention are brain regions responsible for processing social cues and encoding human social behaviour, including the orbital frontal and middle frontal gyrus, superior temporal gyrus, temporal horn, amygdala, anterior precuneus lobe, temporoparietal junction, anterior cingulate cortex, and insula~\cite{nummenmaa2009,akiyama2007unilateral}. From a developmental perspective, infants' attention to social cues helps them quickly learn how to interact with others, learn a language, and build social relationships~\cite{sperdin2018}. {\color{black}However, dysfunctional social attention is one of the primary social impairments for children with Autism Spectrum Disorder (ASD)~\cite{srinivasan2016effects}.} For example, infants with (ASD) are born with less attention to social cues, an inability to track the sight of others, and a fear of looking directly at human faces~\cite{senju2009atypical}. {\color{black}This might be a crucial mechanism that results in their failure to understand others' intentions and engage in typical social interactions~\cite{srinivasan2016effects}.} Research on developmental mechanisms of social attention is still in its early stages. Exploring these scientific questions will be significant for understanding mechanisms of interpersonal social behaviour and developing clinical interventions to assist individuals diagnosed with ASD.

\subsection{Eye Gaze as Social Cue}  
One of the most critical manifestations of social attention is the ability to follow others' eye gazes and respond accordingly~\cite{shepherd2010following}. Eye gaze is proven to have higher social saliency and prioritisation than other social cues~\cite{langton2000eyes} since it indicates to a person the direction in which another person is looking~\cite{frischen2007gaze}. {Gaze following is \color{black}considered as} the foundation of more sophisticated social and cognitive functions like the theory of mind, social interaction, and survival strategies formed by evolution~\cite{baron1997mindblindness,langton2000eyes}. {\color{black}For instance, infants can track the eye gaze of their parents at the age of 3 months~\cite{johnson1998whose,farroni2004gaze,jessen2014unconscious}. After 10 months, gaze following ability significantly contributes to their language development~\cite{brooks2005development,shepherd2010following}.} Psychological studies use the modified Posner cueing task~\cite{posner1984components} or named gaze-cueing task~\cite{friesen1998eyes} to study reflexive attentional orienting generated by the eye gaze. During the task, the eye gaze is presented as the visual cue in the middle of the screen, followed by a peripheral target, which could be spatially congruent (e.g., a right-shift eye gaze followed by a square frame or a Gabor patch shown on the right side of the screen), or incongruent. However, studying the visual modality alone is not enough to reveal how humans can quickly recognise social and emotional information conveyed by others in an environment full of multimodal information~\cite{battich2020coordinating}. Selecting information from the environment across different sensory modalities allows humans to detect crucial information such as life threats, survival strategies, etc.~\cite{newport2009short,fu2020can}. Therefore, several studies conducting a crossmodal gaze-cueing task demonstrate the reflexive attentional effect of the visual cue on the auditory target~\cite{doruk2018cross,maddox2014directing}. Most of these studies rely on images of gaze shifts as visual cues to trigger the observers' social attention~\cite{newport2009short,nuku2010one}. However, these images are not dynamic and lack ecological validity.

\subsection{\color{black}{Stimulus-Response-Compatibility tasks and effects} }
{\color{black}
Researchers study humans’ cognitive control mechanism by using the Stimulus-Response-Compatibility (SRC) tasks to measure the behavioral performance and neural activation on conflict processing. The SRC effect measured by those tasks reflects humans’ better performance in the Stimulus-Response congruent conditions than the incongruent conditions. The classic SRC tasks conducted in the lab are Stroop task~\cite{stroop1935studies}, Flanker task~\cite{eriksen1974effects}, and Simon task~\cite{simon1967auditory}. The size of the SRC effect represents the capacity of conflict processing. The larger SRC effect may be accompanied with the weaker top-down control, dysfunction or immaturity of conflict control~\cite{cohen1990control,mcneely2003neurophysiological}.}

\subsection{\color{black}{Audio-Visual Saliency Modelling}} 
{\color{black}
Saliency prediction models are trained on eye tracking data collected from multiple participants looking at images or videos under the free-viewing condition.
Several studies show that audio-visual input improves models' performances in predicting saliency. Tavakoli et al.~\cite{tavakoli2020} propose a late fusion audio-visual model for enhancing saliency prediction compared to visual-only models. Tsiami et al.~\cite{tsiami2020stavis} show that the early fusion of auditory and visual stimuli reduces reliance on visual content when inferring salient regions. Jain et al.~\cite{jain2020vinet} compare multiple approaches for integrating the two modalities within different layers of the model hierarchies. In contrast to previous findings, the authors show that auditory input degrades performance, suggesting that better audio-visual integration methods are needed. Moreover, sound localisation performances of monaural audio-visual models cannot surpass binaural audio-visual models~\cite{wu2021binaural,rachavarapu2021localize}. This is due to the reduced ability of monaural models to accurately localise sound since the interaural temporal and level difference cannot be computed~\cite{wightman1997monaural}.
Since our task relies mainly on sound direction, we design a binaural sound localisation model that infers saliency both from auditory and visual stimuli.
\newline
\newline
\newline
\newline
}

{\color{black}
\section{Human Experiment}
\label{sec:human_experiment}}

\subsection{Participants}
 37 participants (female = 20) participated in this experiment. Participants were between 18 to 29 years of age, with a mean age of 22.89 years. All participants reported no history of neurological conditions (seizures, epilepsy, stroke, etc.) and had either normal or corrected-to-normal vision and hearing. This study was conducted following the principles expressed in the Declaration of Helsinki. Each participant signed a consent form approved by the Ethics Committee of the Institute of Psychology, Chinese Academy of Sciences.

  \begin{figure*}[ht!]
    \centering
    \subfloat[]{
        \includegraphics[height=0.23\textheight]{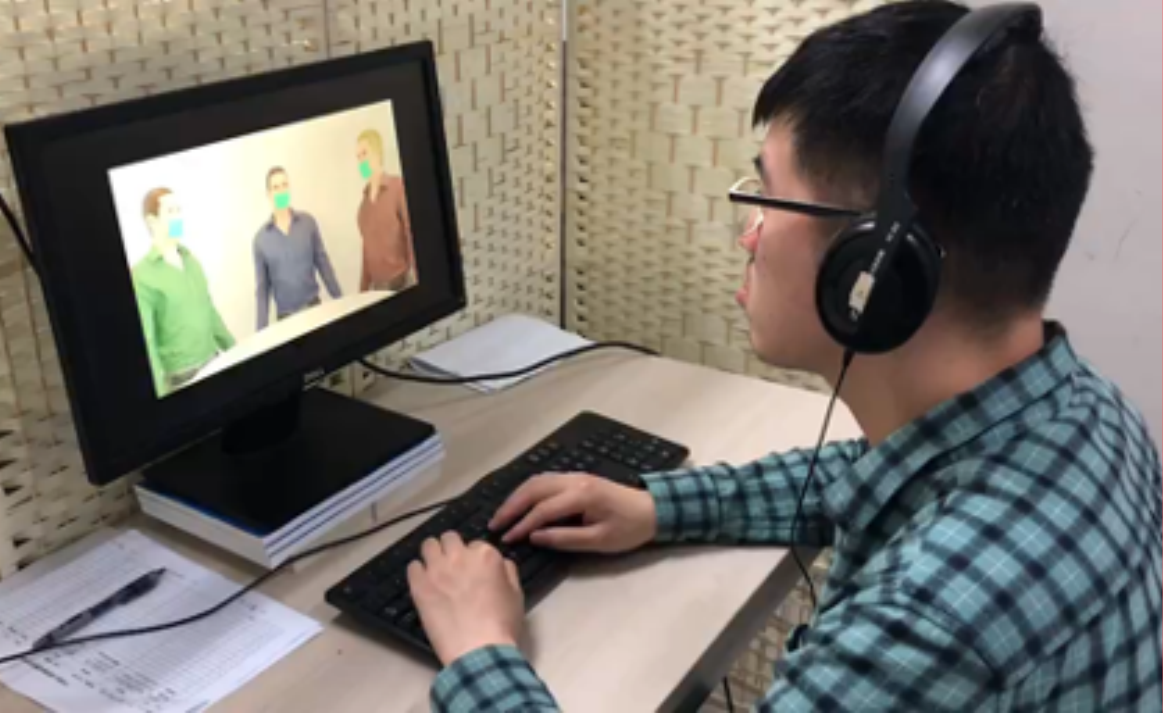}
        \label{fig:av_social_attention_task_human}
    }\hfil
    \subfloat[]{
        \includegraphics[height=0.23\textheight]{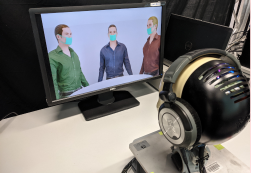}
        \label{fig:av_social_attention_task_robot}
    }
    
    \subfloat[]{
        \includegraphics[height=0.215\textheight]{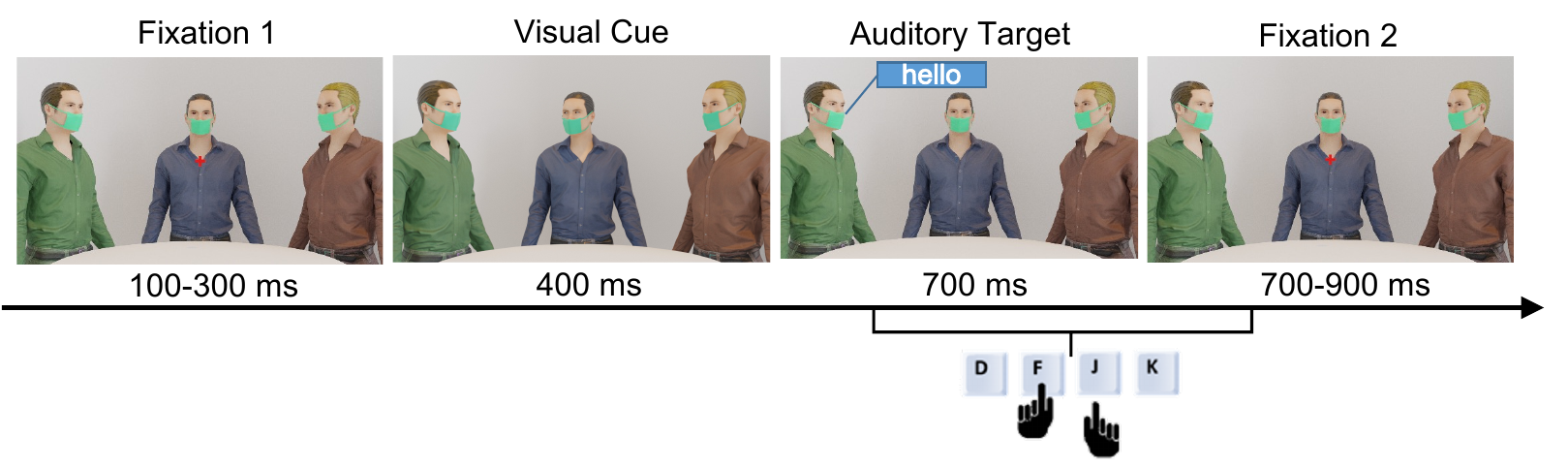}
        \label{fig:av_social_attention_task_trial}
    }
    \caption{Audio-visual gaze-cueing social attention task. \textbf{a.} Participant engaging in the formal test with a headphone to hear the auditory stimulus, and a keyboard to respond; \textbf{b.} The iCub robot engaging in the test with a headphone to get the auditory input, and responding to the target by moving its eyeballs (see Figure~\ref{fig:icub_head}); \textbf{c.} Schematic illustration of a single trial.}
    \label{fig:av_social_attention_task}
\end{figure*}

\subsection{Experimental Setup}
{\color{black}All participants watch clips under normal indoor light conditions. Auditory noise in their surroundings is minimal, and the room acoustic effects are negligible since the sound is played directly through on-ear headphones. This section describes the stimuli generation procedure, the environmental setup, and the data recording methodology.}
\subsubsection{Apparatus, Stimuli and Procedure}
\label{sec:human_apparatus}
Virtual avatars are chosen over recordings of real people, as the experiment requires strict control over the avatar's behaviour, both in terms of timing and exact motion. By using synthetic data as the experimental stimuli, it can be ensured, for instance, that looking to the left and right are exactly symmetrical motions, thus avoiding any possible bias. Moreover, using three identical avatars that are only different in terms of clothing colour also alleviates a bias towards individuals in a real setting. The static basis for the highly-realistic virtual avatars was created in MakeHuman\footnote{\url{http://www.makehumancommunity.org/}}. Based on these avatar models, a data generation framework for research on shared perception and social cue learning with virtual avatars~\cite{kerzel2020} (realised in Blender\footnote{\url{https://www.blender.org/}} and Python) {\color{black}is used to create} the animated scenes with the avatars, which are used as the experimental stimuli in this study. The localised sounds are created from a single sound file using a head-related transfer function\footnote{\url{https://sound.media.mit.edu/resources/KEMAR.html}} that modifies the left and right audio channels to simulate different latencies and damping effects for sounds arriving from different directions. In our 3-avatar scenario, the directions are frontal left and frontal right at 60 degrees, corresponding to the positions where the peripheral avatars stand. 

During the experiment, the participants sit positioned 55 cm from the monitor at a desk and wear headphones, as depicted in Figure~\ref{fig:av_social_attention_task_human}. In each trial, a fixation cross appears in the middle of the screen for 100-300 ms with equal probability. Next, a visual cue is displayed for 400 ms, consisting of an eye gaze shift and a synchronised slight head and upper body shift from the central avatar. In each trial, the central avatar randomly chooses to look at the avatar at the right, at the one at the left, or directly towards the participant, meaning no eye gaze shift at all. Afterwards, the left or the right avatar says ``hello'' with a human male voice as the auditory target. This step lasts for 700 ms. Finally, another fixation cross is shown at the centre of the screen for 700, 800 or 900 ms, with equal probability, until the end of the trial (cf. Figure~\ref{fig:av_social_attention_task_trial} for a schematic representation of the trial).

{\color{black}The experimental design has three directions for the visual cue (left, right, and central) and two for the auditory target location (left, right). The congruent audio-visual condition occurs when the central avatar's eye gazes in the same direction as the avatar who generates the sound. The incongruent audio-visual condition occurs when the central avatar's eye gazes in the opposite direction as the avatar who generates the sound.
The neutral condition is when the central avatar does not shift its eye gaze, so there is no spatial conflict between the visual cue and the following auditory target.
The participants begin the experiment with 30 practice trials and enter into the formal test when their accuracy of practice trials reaches 90\%. Each condition is repeated 96 times, with a total of 288 trials separated into four blocks. There is a 1-minute rest between every two blocks. The time duration for each trial is 1900-2300 ms, and the formal test lasts for 12 minutes.}

{\color{black}During the task, the participants are asked to determine as soon and precisely as possible whether the auditory stimulus originated from the avatar on the left or on the right. The participants make decisions by pressing the keys ``F'' and ``J'' on the keyboard, corresponding to the left and right avatars. The participants' responses during the display of the auditory target and the second fixation are recorded. The stimulus display and response recording are both under the control of E-prime 2.0\footnote{Schneider, W., Eschman, A., \& Zuccolotto, A. (2002). E-Prime (Version 2.0). [Computer software and manual]. Pittsburgh, PA: Psychology Software Tools Inc.}. In the current study, all participants perceive the simulated masks as typical.}

\subsubsection{Data Recording and Analyses} 
Reaction time (RT) and error rates (ER) are analysed as human response indices. For the RT analysis, error trials, and trials with RTs shorter than 200 ms, and those with RTs beyond three standard deviations above or below the mean were excluded, corresponding to 2.42\% of the data being removed. To examine the Stimulus-Response Compatibility effects of the crossmodal audio-visual conflict task, one-way repeated measures analysis of variance (ANOVA) is used to test differences in the participants' responses under the three congruency conditions (congruent, incongruent and neutral). All post hoc tests in the current study use Bonferroni correction.

\subsection{\color{black}{Experimental Results}}
{\color{black}Our experimental results indicate that the participant response time and accuracy under the audio-visually congruent condition exceeded the performance under the incongruent condition. {\color{black}There are no significant differences between the neutral and incongruent conditions for both RT and ER. The lack of difference between the neutral and incongruent conditions shows that the lack of congruent audio-visual cueing negatively affects the participants' performance.} }
\subsubsection{Reaction Time}
\begin{figure*}[ht!]
    \centering
    \subfloat[]{
        \includegraphics[height=0.2\textheight]{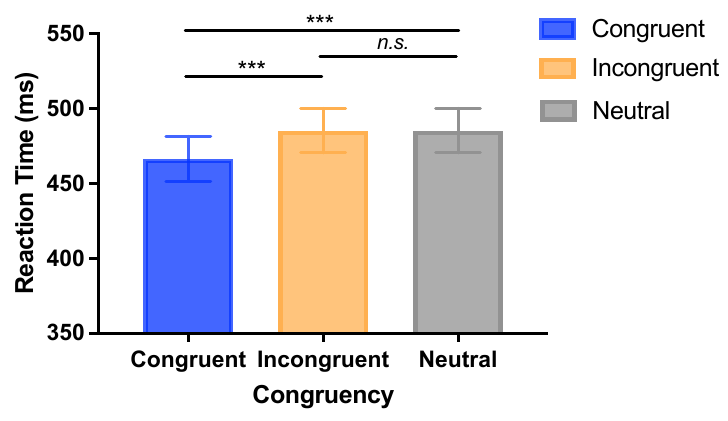}
        \label{fig:rt_human_group}
    }\hfil
    \subfloat[]{
        \includegraphics[height=0.2\textheight]{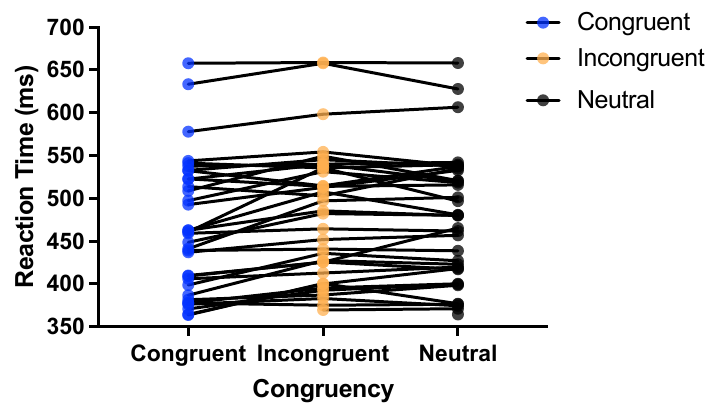}
        \label{fig:rt_human_individual}
    }
    
    \subfloat[]{
        \includegraphics[height=0.2\textheight]{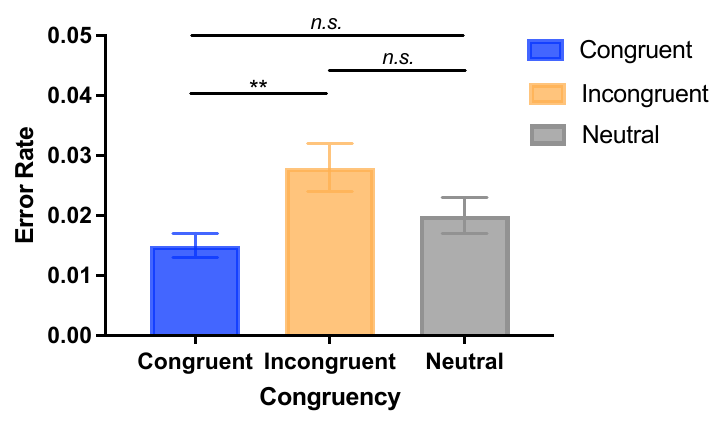}
        \label{fig:er_human_group}
    }\hfil
    \subfloat[]{
        \includegraphics[height=0.2\textheight]{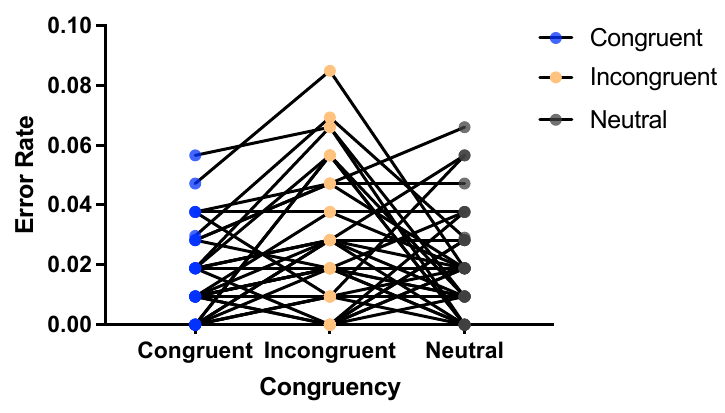}
        \label{fig:er_human_individual}
    }
    
    \subfloat[]{
        \includegraphics[height=0.2\textheight]{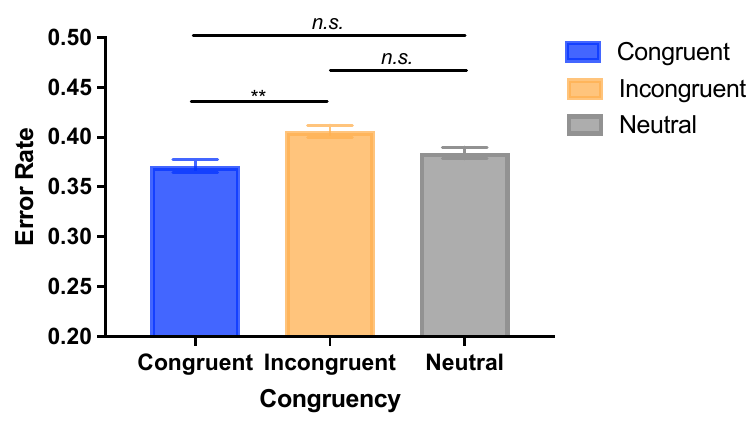}
        \label{fig:er_robot_group}
    }\hfil
    \subfloat[]{
        \includegraphics[height=0.2075\textheight]{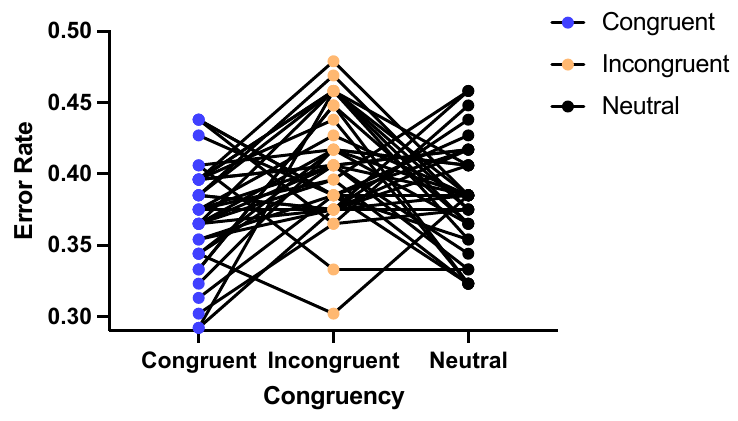}
        \label{fig:er_robot_individual}
    }
    \caption{\textbf{a.} RT of participants under different congruency conditions -- group level; \textbf{b.} RT of participants under different congruency conditions -- individual level; \textbf{c.} ER of participants under different congruency conditions -- group level; \textbf{d.} ER of participants under different congruency conditions -- individual level; \textbf{e.} ER of the iCub under different congruency conditions -- group level; \textbf{f.} ER of the iCub under different congruency conditions -- individual level. $*$ denotes $.01 < p < .05$, $*\!*$ $.001 < p < .01$, $*\!*\!*$ $p < .001$, and \textit{n.s.} denotes no significance.}
    \label{fig:analysis_graphs}
\end{figure*}

A repeated measures ANOVA with a Greenhouse-Geisser correction shows that the participants' RT differs significantly between different congruency conditions, $F \left( 2 , 34 \right) = 24.19, p < .001, \eta_{p}^{2} = .40$ (see Figures~\ref{fig:rt_human_group} and~\ref{fig:rt_human_individual}). Post hoc tests show that the participants responded significantly faster under the congruent condition ($\text{mean} \, \pm \, \text{SE} = 466.25 \pm 14.92 \, \text{ms}$) than both incongruent condition ($\text{mean} \, \pm \, \text{SE} = 485.12 \pm 14.82 \, \text{ms}, p < .001$) and neutral condition ($\text{mean} \, \pm \, \text{SE} = 485.11 \pm 14.80 \, \text{ms}, p < .001$). However, the difference between the incongruent and neutral condition was not significant, $p > .05$.

\subsubsection{Error Rates}
 A repeated measures ANOVA with a Greenhouse-Geisser correction shows that the participants' ER differs significantly between different congruency conditions, $F \left( 2 , 34 \right) = 5.69, p < .05, \eta_{p}^{2} = .14$ (see Figures~\ref{fig:er_human_group} and~\ref{fig:er_human_individual}). Post hoc tests show that the participants presented significantly lower ER under the congruent condition ($\text{mean} \pm \text{SE} = .02 \pm .002$) than the incongruent condition ($\text{mean} \pm \text{SE} = .03 \pm .004$), $p < .01$. However, there was no statistical significance in the difference between the neutral condition ($\text{mean} \pm \text{SE} = .02 \pm .003$) and both other congruency conditions, $p > .05$ in both cases.

\begin{figure*}[ht!]
    \centering
    \subfloat[]{
        \includegraphics[height=0.215\textheight]{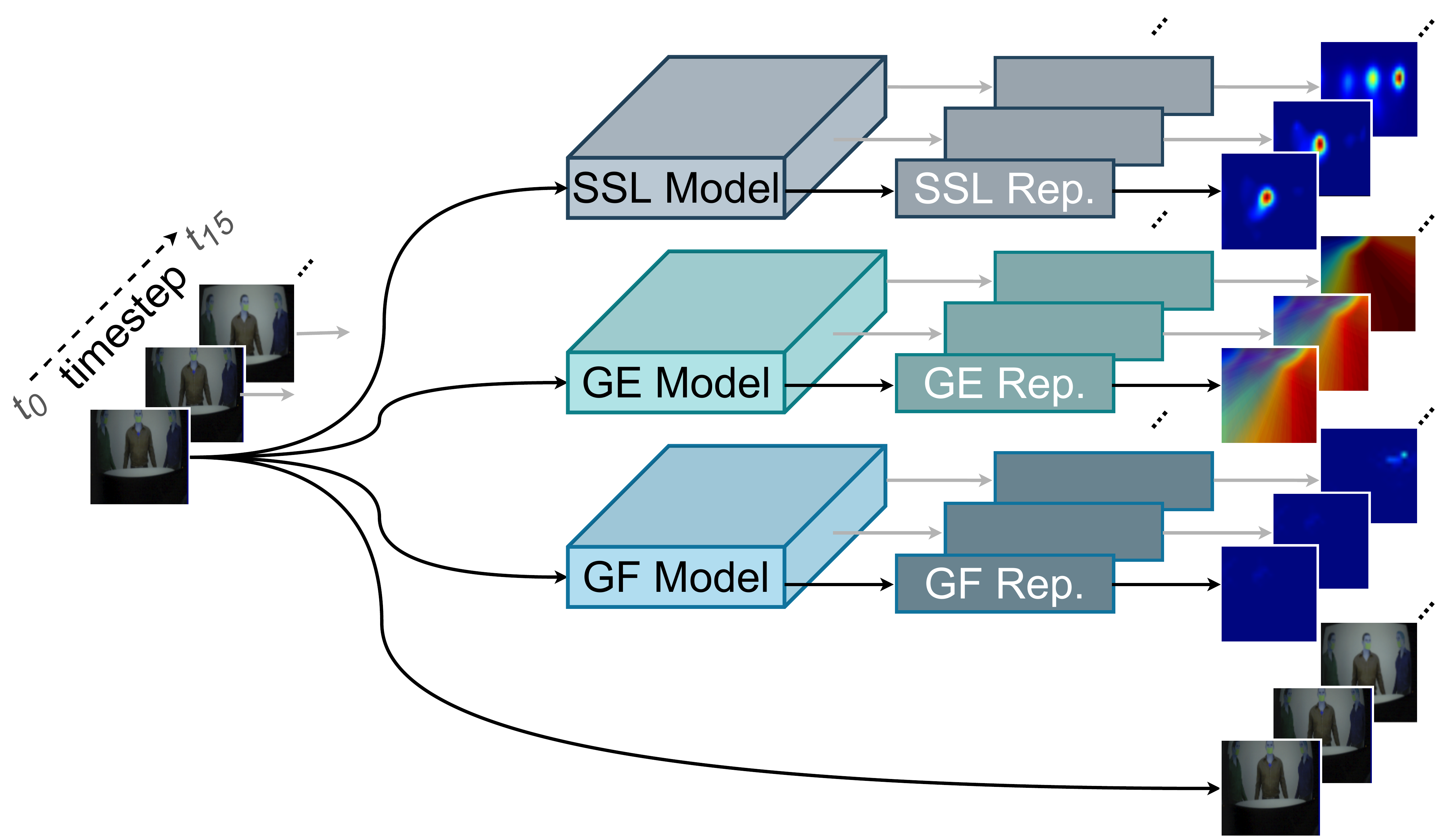}
        \label{fig:scd}
    }
    \subfloat[]{
        \includegraphics[height=0.23\textheight]{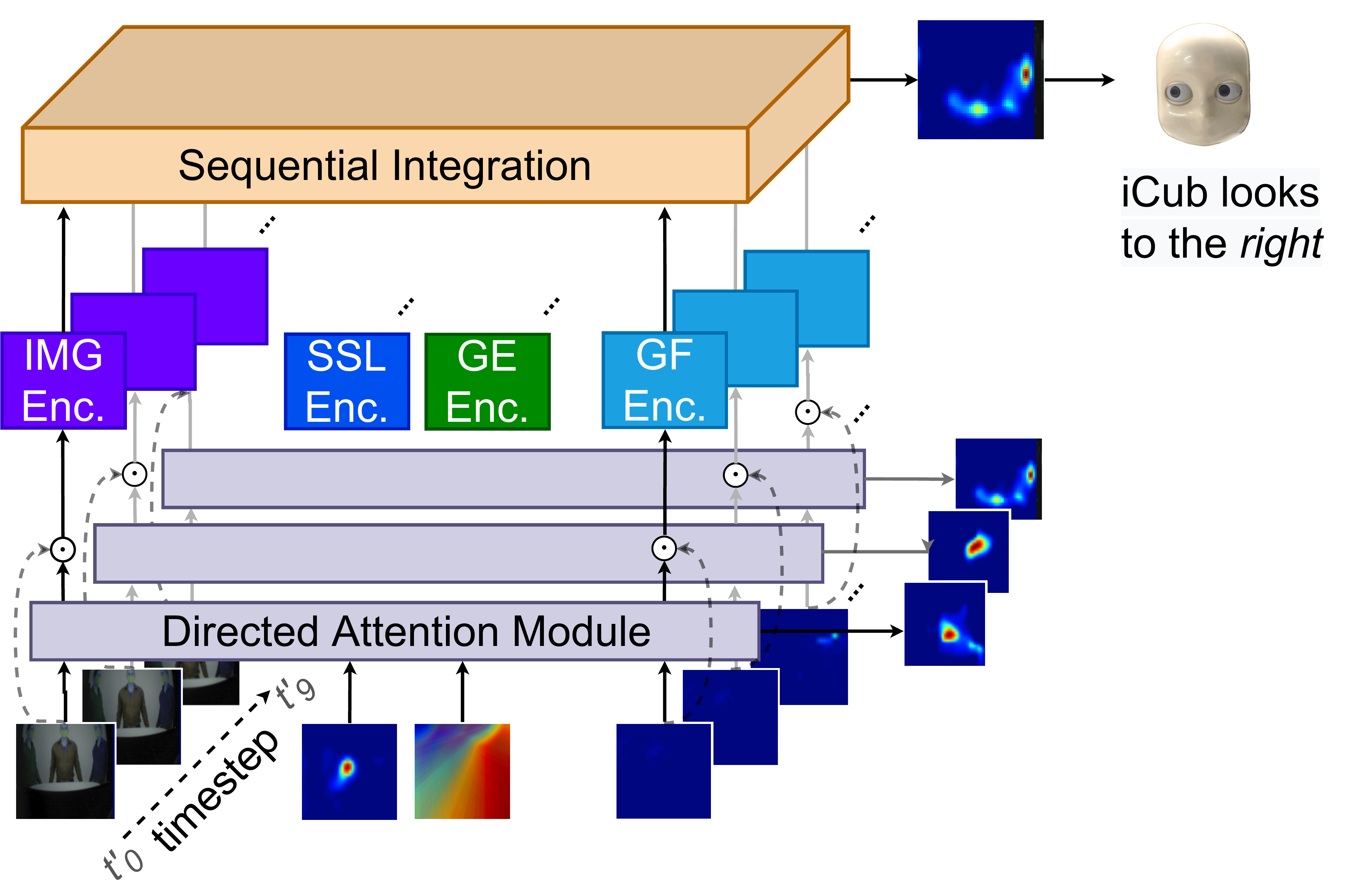}
        \label{fig:gasp}
    }
    
    \subfloat[]{
        \includegraphics[height=0.27\textheight]{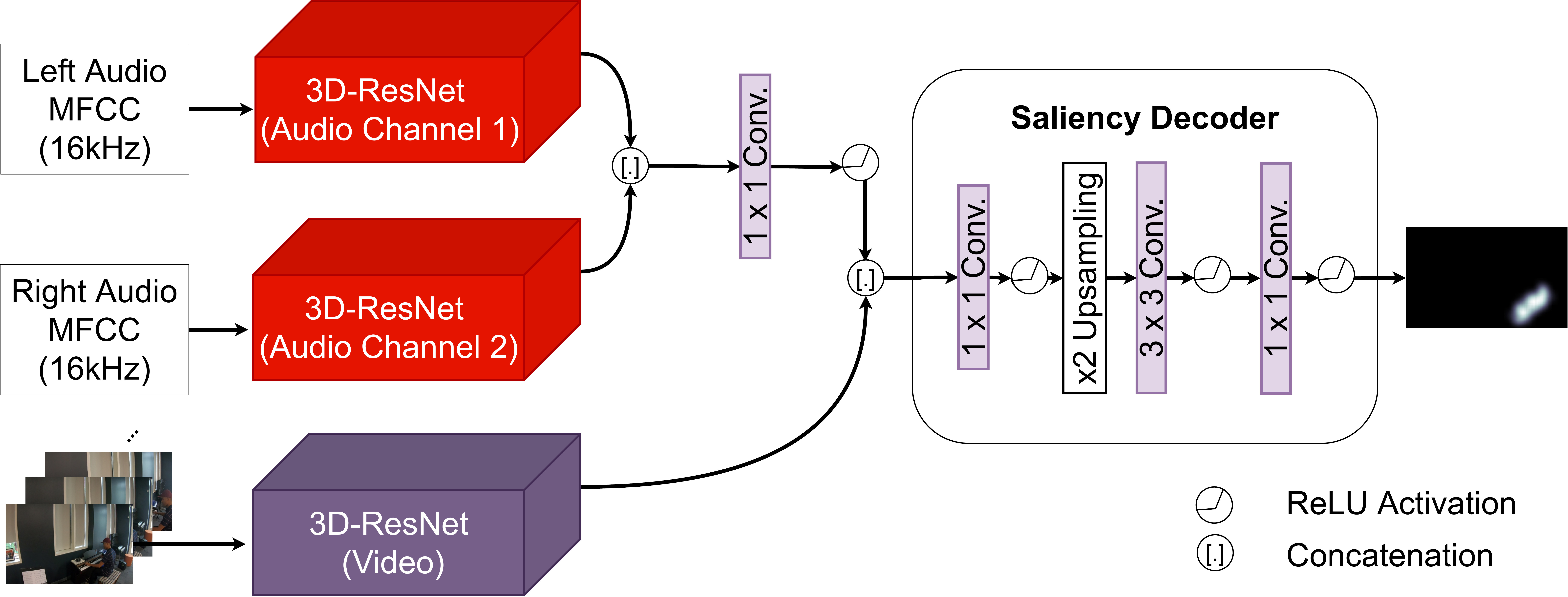}
        \label{fig:binaural_dave}
    }
    \caption{\textbf{a.} SCD -- Social cue detection stage in which the representations of the sound source localisation (SSL), gaze estimation (GE), and gaze following (GF) are extracted; \textbf{b.} GASP -- Saliency prediction; \textbf{c.} Binaural DAVE -- Audio-visual sound source localisation}
    \label{fig:gasp_structures}
\end{figure*}

\color{black}\section{Robot Experiment}
\label{sec:gasp}

\begin{figure}[ht!]
    \centering
    \includegraphics[width=0.425\textwidth]{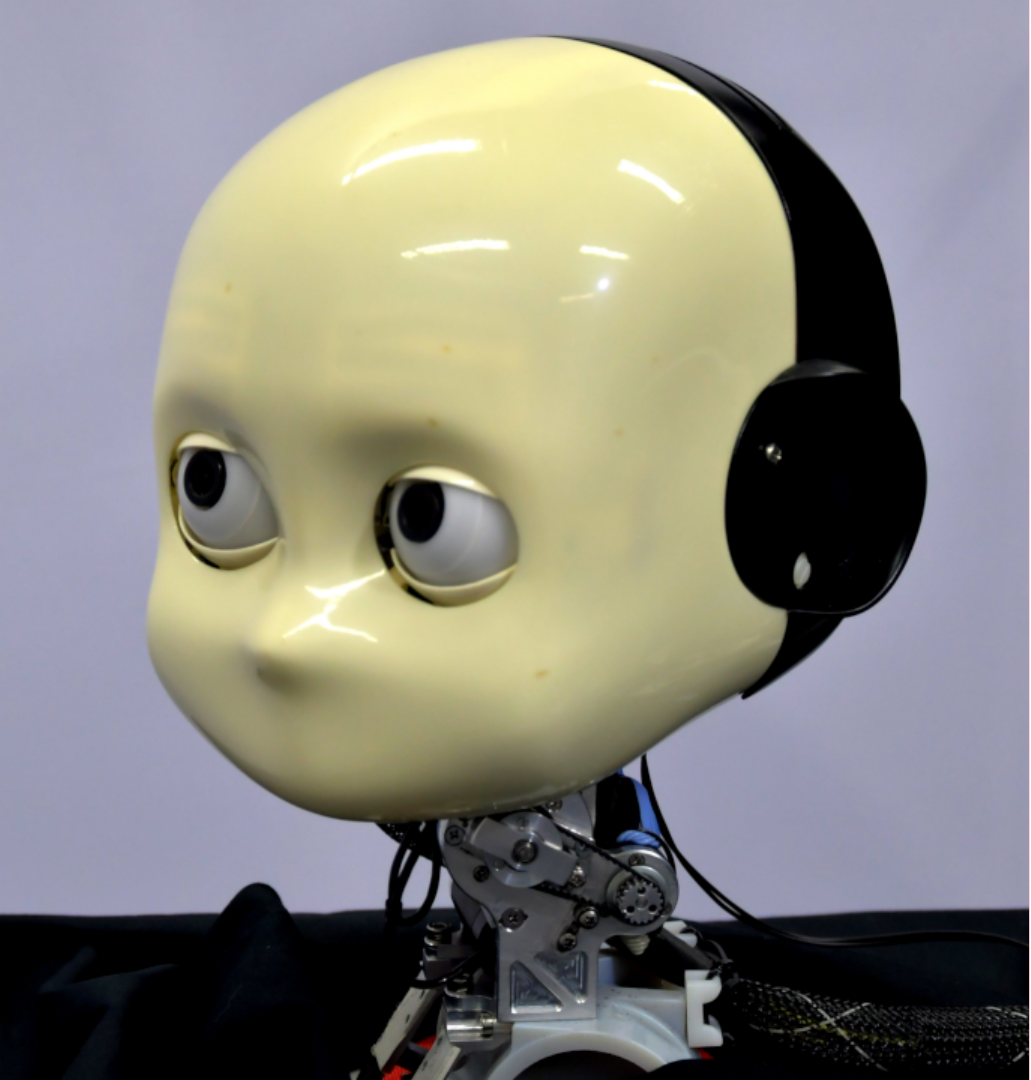}
    \caption{The iCub head.}
    \label{fig:icub_head}
\end{figure}

\subsection{\color{black}{Neural Modelling}}
{\color{black}We aim to assess whether the iCub head would display degradation in performance under the incongruent condition relative to the congruent condition. {\color{black}To achieve that, we require a model capable of dealing with stimuli from the gaze following modality showing the attention targets of all individuals observed in the video, as well as the gaze estimation modality, indicating their head and eye poses, along with audio source localisation}. For that purpose, we opted for using the GASP model~\cite{abawi2021}, an audio-visual saliency predictor receiving the aforementioned social cues as input. However, GASP was originally designed to work solely with monaural inputs. Since the auditory stimulus in the three-avatar scenario arrives from a single direction, we modify GASP to accommodate stereo audio. We do so by replacing the saliency prediction model with a binaural sound localisation model.}

\subsubsection{\color{black}{Dynamic Saliency Prediction}}
The process of predicting saliency is divided into two stages. The first stage, Social Cue Detection (SCD), is responsible for extracting social cue feature maps from a given audio-visual sequence. Figure~\ref{fig:scd} depicts the architecture of the SCD stage. Given a sequence of images and their corresponding high-level feature maps, the second stage, GASP, then predicts the corresponding saliency region by integrating the social cue feature map sequences. The overall integration pipeline followed by GASP is shown in Figure~\ref{fig:gasp}.
        
Following the implementation of GASP, the SCD stage comprises four modules, each responsible for extracting a specific social cue~\cite{abawi2021}. Those modules include gaze following, gaze estimation, facial expression recognition, and audio-visual saliency prediction. For the current task, however, the facial expression recognition module is not employed since the virtual avatar faces are partially occluded and do not display facial expressions. In order to closely replicate the experiments done with participants, the iCub robot receives auditory stimuli from both ears. An audio-visual saliency prediction module was originally designed to work with monaural stimuli. To operate on binaural stimuli, we replace the saliency prediction module with a binaural audio-visual sound source localisation (SSL) model, denoted the ``SSL model'' in Figure~\ref{fig:scd}. The binaural SSL model architecture is {\color{black}shown} in Figure~\ref{fig:binaural_dave}.
        
The video streams used as input are split into their frames and corresponding auditory signals. For every video frame and corresponding audio signal, the SCD stage covers the extraction of social cue feature maps, which are then propagated to GASP. The Directed Attention Module (DAM) weighs the feature map channels to emphasise those that represent high unexpectedness with respect to their predictions. Convolutional layers further encode those weighted feature map channels. In Figure~\ref{fig:gasp}, these layers are denoted by ``Enc.'' (for encoder). The encoded feature maps of all video frames are then integrated using a recurrent extension of the convolutional Gated Multimodal Unit (GMU)~\cite{arevalo2020}. The GMU's mechanism weighs the features of its inputs. Adding a convolutional aspect to it accounts for the preservation of spatial properties of the input features. The recurrent property of the integration unit considers the whole sequence of frames by performing the gated integration at every timestep.
        
For this work, the LARGMU (Late Attentive Recurrent Gated Multimodal Unit) is used because of its better performance compared to other GMU-based models~\cite{abawi2021}. Since LARGMU is based on the convolution GMU, it preserves the input spatial features. The LARGMU's recurrent structure allows it to integrate those features sequentially. Adding a soft-attention mechanism based on the convolutional Attentive Long Short-Term Memory (ALSTM)~\cite{cornia2018} prevents gradients from vanishing as feature sequences get sufficiently large. As the name implies, LARGMU is a late fusion unit, meaning that the gated integration is performed after the input channels are concatenated and, in sequence, propagated to the ALSTM.
    
\subsubsection{\color{black}{Binaural Sound Localisation}}
    
{\color{black}DAVE (Deep Audio-Visual Embedding)~\cite{tavakoli2020} is used as a sound source localisation module in the SCD stage. In its original form, the audio-visual DAVE encodes inputs from one video and one audio stream, which are projected into a feature space by 3D-ResNets~\cite{hara2018can} (one for each input stream). {\color{black}3D-ResNet extends the ResNet model ~\cite{he2016identity} to operate on multiple frames by replacing 2D convolutional layers with their 3D counterparts.} Its encoder is followed by a convolutional saliency decoder that upscales the latent representation and provides the corresponding saliency map. For our current work, DAVE is extended to accept binaural input, and this binaural extension structure uses a similar rationale to the monaural DAVE, see Figure ~\ref{fig:binaural_dave}. The main difference is using two 3D-ResNets to process the auditory modality, whose output features are concatenated and then encoded and downsampled by a two-dimensional convolutional layer. This layer is responsible for guaranteeing that the dimension of the feature produced by this part of the architecture matches that of the feature produced by DAVE's original audio-stream 3D-ResNet.}
        
{\color{black}We initialise the binaural DAVE with the pre-trained parameters of the audio-visual DAVE~\cite{tavakoli2020}. The left and right auditory streams are initialised with identical parameter weights extracted from the 3D-ResNet auditory stream of the monaural variant. The $1 \times 1$ convolutional layer that encodes the concatenated audio features is initialised using the normalisation method proposed by He et al.~\cite{he2015}. All model parameters are optimised except for the video 3D-ResNet, which are frozen throughout optimisation following DAVE's training procedure~\cite{tavakoli2020}.}

\subsubsection{\color{black}{Binaural DAVE as a Prior to GASP}}
        
The GASP architecture used in the experimental setup consists of the pretrained GASP, excluding the facial expression recognition input stream. We replace the audio-visual saliency detector with DAVE's binaural sound localisation variant. Abawi et al.~\cite{abawi2021} show that replacing saliency predictors does not require re-training GASP, allowing us to use a sound localisation model in the place of a saliency prediction model without fine-tuning the sequential integration parameters.
        
GASP receives four sequences of data as input, one sequence of consecutive frames of the original video and three sequences of feature maps, one for each model in the social cue detection stage. In our experiment, we capture sequences of 10 frames (cf. timesteps $t^{\prime}_{0}$ to $t^{\prime}_{9}$ in Figure~\ref{fig:gasp}). The number of frames received as input by each model in the SCD stage varies due to dissimilarity in their expected inputs. The sound localisation model receives a sequence of 16 frames as input, whereas the gaze estimation and following models receive sequences of 7 frames each. A more detailed explanation of how the frames are selected based on the timestep being processed is provided by Abawi et al.~\cite{abawi2021}. The auditory input is captured as a one-second chunk and propagated to each audio 3D-ResNet of the sound localisation model. In this experiment, GASP is embedded in the iCub robot and subjected to the same series of one-second videos as the participants. The one-second chunk used as input to the binaural sound localisation model corresponds to the entire audio recording per video.

\subsection{iCub Eye Movement Determination}
    
After the iCub acquires the visual and auditory inputs, the social cue detectors and the sound source localisation model extract features from those audio-visual frames. Following the detection and generation of the feature maps, they are propagated to GASP, which, in turn, predicts a fixation density map $\mathcal{F} \colon \mathbb{Z}^2 \to \left[ 0 , 1 \right]$, which is displayed in the form of a saliency map for a given frame. The fixation peak $\left( x_{\mathcal{F}} , y_{\mathcal{F}} \right)$ is determined by calculating
\begin{equation}
    \left( x_{\mathcal{F}} , y_{\mathcal{F}} \right) = \mathop{\mathrm{argmax}}\nolimits_{x , y} \mathcal{F} \left( x , y \right).
    \label{eq:fdm_peak}
\end{equation}
The values of $x_{\mathcal{F}}$ and $y_{\mathcal{F}}$, originally in pixels, are then normalised to scalar values $\hat{x}_{\mathcal{F}}$ and $\hat{y}_{\mathcal{F}}$ within the $\left[ -1 , 1 \right]$ range, such that
\begin{align}
    \hat{x}_{\mathcal{F}} &= \dfrac{2 x_{\mathcal{F}}}{l_{x}} - 1, \\
     \hat{y}_{\mathcal{F}} &= \dfrac{2 y_{\mathcal{F}}}{l_{y}} - 1, \label{eq:fdm_peak_xy_norm}
\end{align}
where the width $l_{x}$ and height $l_{y}$ indicate the number of fixation density map pixels in each axis. A value of $\hat{x}_{\mathcal{F}}  = -1$ represents the left-most point and $\hat{x}_{\mathcal{F}} = 1$ the right-most one. The vertical axis, $\hat{y}_{\mathcal{F}} = -1$ represents the top-most point and $\hat{y}_{\mathcal{F}} = 1$ the bottom-most one.
        
The robot is actuated to look towards the fixation peak. For simplicity, eye movement is assumed to be independent of the exact camera location relative to the playback monitor. For all experiments, only the iCub eyes were actuated while disregarding microsaccadic movements and vergence effects. The positions the iCub should look at are expressed in Cartesian coordinates while assuming the monitor to be at a distance of 30 cm ($\delta = 0.3$) from the image plane. To limit the viewing range of the eyes, $\hat{x}_{\mathcal{F}}$ and $\hat{y}_{\mathcal{F}}$ are scaled down by a factor of $\alpha = 0.3$. The Cartesian coordinates are then converted to spherical coordinates by
\begin{align}
    \theta &= \arctan \left( \dfrac{\alpha \cdot \hat{x}_{\mathcal{F}}}{ \sqrt{\delta^{2} + (\alpha \cdot\hat{y}_{\mathcal{F}})^{2}}} \right), \\
    \phi &= \arctan (\hat{y}_{\mathcal{F}}), \label{eq:yaw_pitch}
\end{align}
where $\theta$ and $\phi$ are the yaw and pitch angles respectively. These angles are used to actuate the eyes of the iCub such that they pan $\sim\!27^{\circ}$ and tilt $\sim\!24^{\circ}$ at most\footnote{The iCub can pan its eyes within a $\left[ -45^{\circ} , 45^{\circ} \right]$ range and tilt them within a $\left[ -40^{\circ} , 40^{\circ} \right]$ range.}.

\subsection{Experimental Setup}
{\color{black}We train the binaural model on a stereo audio-visual dataset and propagate its predicted maps to GASP. We describe the physical setup of the robot environment under which the model used for integrating social cues with binaural sound is evaluated. The human and robot experimental setups closely resemble each other, allowing us to emulate the environmental surrounding experienced by the participants that was described in Section~\ref{sec:human_apparatus}.}

\subsubsection{Binaural Model Training and Evaluation}

{\color{black}The binaural DAVE is fine-tuned on a subset of the FAIR-Play dataset~\cite{gao2019}, comprising 500 randomly chosen videos. The FAIR-Play dataset consists of 1,871 video clips of single or multiple individuals playing musical instruments indoors. Auditory input is binaural with the sound source location maps provided by Wu et al.~\cite{wu2021binaural}.

{\color{black}Similar to its monaural counterpart, the loss of the binaural DAVE model is computed as Kullback-Leibler divergence between the predicted and ground-truth fixation maps at the last timestep of the 16-frame sequence. The input frames, sound channels and ground-truth maps were together flipped at random during training as an augmentation transform. We use the Adam optimiser with $\beta_1=.9$, $\beta_2=.999$, and a learning rate of $.001$. The model is trained for five epochs with mini-batches containing four sequences of 16 visual frames with their corresponding one-second stereo recordings of audio. We train the model on an NVIDIA GeForce RTX 3080 Ti with 32 GB RAM.}

{\color{black}We test our model on 200 randomly chosen clips from the FAIR-Play dataset. Another set of 200 clips are used for validation. Given the close resemblance of audio-visual sound localisation to saliency modelling, we rely on metrics commonly used to evaluate the latter~\cite{bylinskii2019eval}. We measure Pearson's correlation coefficient (CC) and similarity (SIM) to quantify the performance of our model. CC calculates the linear correlation between two normalised variables, whereas SIM signifies the similarity between two distributions with a value of 1 indicating that they are identical.} 

\subsubsection{Physical Robot Environment}
\label{sec:icub}
Some technical adjustments proved necessary to replicate the human experiments on the iCub head as closely as possible. First, the iCub head was placed at a distance of approximately 30 cm from a 24-inch monitor ($1920 \times 1200$ pixel resolution), as depicted in Figure~\ref{fig:av_social_attention_task_robot}. This distance is, however, shorter than the 55 cm distance the participants sat from the desktop screen. The distance reduction was performed so that the iCub's field of vision covers a larger portion of the monitor. Since the robot lacks foveated vision, the attention is distributed uniformly to all visible regions, causing the robot to attend to irrelevant environmental changes or visual distractors. Second, the previous robot's eye fixation position needed to be retained as a starting point for the next trial to provide scenery variations to the model. Direct light sources also needed to be switched off to avoid glare. Once the experimental setup was ready, the pipeline started the video playback in fullscreen mode, simultaneously capturing a 30-frame segment of the video using a single iCub camera\footnote{\url{http://wiki.icub.org/wiki/Cameras}} along with one-second audio recordings from each microphone\footnote{\url{http://wiki.icub.org/wiki/Microphones}} mounted on the iCub's ears. The video segments were propagated directly from the iCub to the model using the YARP~\cite{yarp2006metta} middleware. Since YARP transmits messages with a low latency and is supported by the iCub, we chose YARP for image transfer. The audio chunks on the other hand were transmitted using the ZeroMQ~\cite{zeromq2013hintjens} middleware  given its lower packet drop rate compared to YARP. Switching between different middleware was facilitated using Wrapyfi~\cite{abawi2023wrapyfi}, a python wrapper for multi-middleware support.
        
In the current study, the iCub head shifts its eyes towards the auditory target. {\color{black}This differs from how participants responded to the stimuli. The participants provided feedback by pressing a key, with their hands were already resting on the keyboard, leading to a much faster response than the time it takes for an iCub head to shift its eyes.} This difference could lead to systematic differences in RT{\color{black}, making the RT of the iCub head incomparable to those of the participants. For that reason,} the RT of the robot was not measured nor analysed. Nevertheless, it is worth noticing that even though humans and the robot respond differently to a trial, the task they perform is essentially the same. Therefore, ER can be adequately measured and analysed as the robot response. One-way repeated measures ANOVA is used to test the SRC effects of the robot's response under the three congruency conditions (congruent, incongruent and neutral). All post hoc tests in the current study use Bonferroni correction. Additionally, an independent $t$-test is conducted to compare the difference in SRC effects between humans and the robot. The SRC effect is measured by subtracting congruent responses from incongruent responses.

\subsection{Experimental Results}
{\color{black}Our binaural audio-visual sound localisation model outperforms monaural and visual-only variants in terms of the CC and SIM metrics.} For processing conflicting auditory and visual stimuli, using a binaural model becomes necessary to estimate the direction of sound arrival. This allows us to replicate human-like patterns in attending to sound under congruent, incongruent, and neutral conditions.

\subsubsection{\color{black}{Binaural Sound Localisation}}
{\color{black}We fine-tune the DAVE variants on the FAIR-Play training subset and evaluate the CC and SIM metrics on the test subset. We compare the predicted saliency maps against the ground-truth audio maps for all video frames. The input consists of the preceding 15 frames of a given video's final frame at timestep $t_{15}$ including the final frame. The evaluation results are reported following the fifth training epoch, given that the validation loss increases after that. The binaural DAVE outperforms both the audio-visual and visual-only variants of DAVE, as shown in Table~\ref{tab:ssl_eval}. 

We observe a significant gap in SIM, but not in CC, between the binaural DAVE and other variants. The SIM metric is highly sensitive to false negatives~\cite{bylinskii2019eval}. Given the objective of localising sounds in the visual stream, saliency prediction models would produce maps uncorrelated with regions having high sound activity. In the case of audio-visual and video-only variants, the models are unaware of the sound location and rely on the activity observed in the visual stream. This implies that those model variants behave like saliency predictors. 

In Figure~\ref{fig:ssl_qual}, we observe that the predictions highly correspond to the ground-truth maps, with an incorrect prediction displayed in the last column. Wrong predictions lead to faulty movement on the iCub during inference. We note that such false predictions often occur due to the labels being provided as constant audio maps for entire video clips~\cite{wu2021binaural}. Changes during the video in which one musician begins playing at a later stage are ignored, as seen from the example shown in the last column of Figure~\ref{fig:ssl_qual}. As indicated by the hand movement in transition between the timesteps $t_0$ and $t_{15}$, the musician is playing the cello.}

\begin{table}[ht!]
\caption{\color{black}{Evaluating the binaural audio-visual sound source localisation model on the test subset of the FAIR-Play dataset.}}
\label{tab:ssl_eval}
\begin{tabular}{lll}
\hline 
Methods                             & CC$\uparrow$ & SIM$\uparrow$ \\ \hline
Visual-only DAVE                    & 0.5030 & 0.3972 \\
Audio-visual DAVE                   & 0.6068 & 0.4398 \\
Binaural audio-visual DAVE (ours)   & \textbf{0.6411} & \textbf{0.5050} \\ \hline
\end{tabular}
\end{table}

\begin{figure}[ht!]
    \centering
    \includegraphics[width=0.425\textwidth]{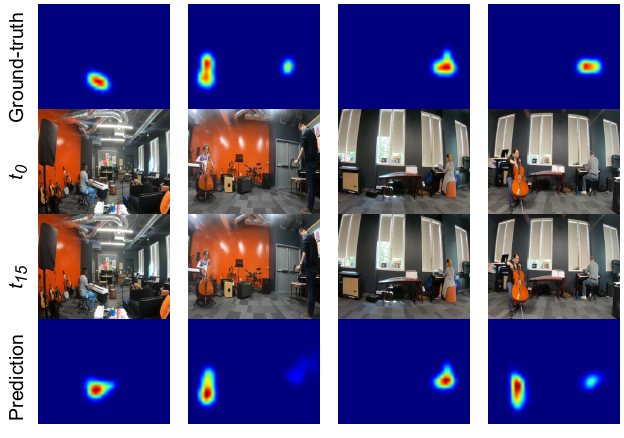}
    \caption{\color{black}Qualitative examples showing the binaural DAVE predictions on the FAIR-Play test subset.}
    \label{fig:ssl_qual}
\end{figure}

\subsubsection{Error Rates}
        
A repeated measures ANOVA with a Greenhouse-Geisser correction showed that the robot's ER differed significantly between different congruency conditions, $F \left( 2 , 34 \right) = 8.02 $, $p < .01$, $\eta_{p}^{2} = .18$ (see Figures~\ref{fig:er_robot_group} and~\ref{fig:er_robot_individual}). Post hoc tests showed that the robot presented significantly lower ER under the congruent condition ($\text{mean} \, \pm \, \text{SE} = .37 \pm .01$) than the incongruent condition ($\text{mean} \, \pm \, \text{SE} = .41 \pm .01 $), $p < .01$. However, there was no statistical significance in the difference between the neutral condition ($\text{mean} \, \pm \, \text{SE} = .38 \pm .01$) and both other congruency conditions, $p > .05$ in both cases.
        
\subsubsection{Human-Robot Comparison}
            
{\color{black}The SRC effect was computed as the difference between ER under incongruent and congruent conditions ($SRC=ER_{incongruent} - ER_{congruent}$).} Results of the $t$-test displayed that the robot showed a significantly larger SRC effect ($\text{mean} \, \pm \, \text{SE} = .04 \pm .001$) than humans ($\text{mean} \, \pm \, \text{SE} = .01 \pm .01$), $t \left( 72 \right) = 2.35$, $p < .05$ (see Figure~\ref{fig:stimulus_response_compatibility_comparison}).

\begin{figure}[!ht]
    \centering
    \includegraphics[width=0.5\textwidth]{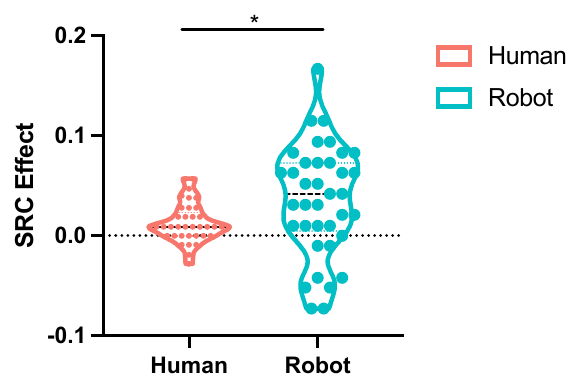}
    \caption{SRC effects comparison between humans and the robot. $*$ denotes $.01 < p < .05$.}
    \label{fig:stimulus_response_compatibility_comparison}
\end{figure}

\section{General Discussion and Conclusions}
\label{sec:discussion}

Our current neurorobotic study investigated human attentional response and modelled the human-like response with the humanoid iCub robot head in an crossmodal audio-visual social attention meeting scenario. According to the research goals, the main findings of the current study are twofold. First, in line with previous crossmodal social attention research~\cite{nuku2008joint,soto2005spatial}, our study shows that the visual cue direction enhances the detection of the following auditory target occurring in the same direction, although from a different modality. The current study uses a dynamic gaze shift with corresponding head and upper body movements as visual cue stimuli. It replicates the previous findings by studies using static eye gaze~\cite{nuku2010one,guo2019abnormal}, showing a robust reflexive attentional orienting effect. More specifically, the participants show longer RT and higher ER under the incongruent audio-visual condition than the congruent one. Some previous research shows that eye gaze has a stronger attentional orienting effect than simple experimental stimuli (e.g., arrows)~\cite{friesen2004attentional,ristic2007attentional}. Although we do not have any conditions using arrows as visual cues in our current study, we first demonstrate that realistic and dynamic social cues could have a similar effect in a human crossmodal social attention behavioural study \textcolor{black}{(\textbf{H1, H2})}. Second, the results from the iCub response demonstrate a successful human-like simulation. With the GASP model, the iCub robot could trigger similar attentional patterns as humans, even in a complex crossmodal scenario. Lastly, the statistical comparison of the SRC effects between humans and the iCub shows that the robot experienced a larger conflict effect than humans \textcolor{black}{(\textbf{H3, H4})}.

In the human experiment, corresponding to our \textcolor{black}{\textbf{H1}} hypothesis, social cues that trigger social attention are extended to multiple modalities. Our results support the nature of social gaze cueing, and the view of stimulus-driven shifts of auditory attention might be mediated by other modality information~\cite{shepherd2010following}. Furthermore, different from previous gaze-cueing experiments~\cite{nuku2008joint}, \textcolor{black}{we add a neutral condition to study the interference and facilitation effects during conflict processing. For the neutral condition, participants only see a static meeting scenario without any dynamic social visual cues before the auditory target comes out.} \textcolor{black}{The results of humans' RT contradict our {\textbf{H2}} hypothesis.} \textcolor{black}{Participants have significantly longer RT under the neutral condition than in the congruent condition. However, no significant difference in RT between neutral and incongruent conditions is found. Thus, the congruent condition in our study has a facilitation effect on the audio-visual conflict processing.
These results are consistent with previous studies using the static eye gaze as the visual cue. Their researchers also report a faster response to the gaze-target spatially congruent conditions than the neutral and incongruent conditions, implying a benefit effect of the gaze-oriented attention~\cite{schuller2004perception,friesen1998eyes}. The ER results show that the incongruent condition would have significantly more response errors than the congruent condition, and the neutral condition intermediates between incongruent and congruent conditions with slight differences.}
    
In the robot experiment, the iCub experiment results verify \textcolor{black}{\textbf{H3}}  and \textcolor{black}{\textbf{H4}} that, similarly to humans, the robot's response accuracy is significantly better ($p < .01$) in a congruent condition than in an incongruent one. This similarity is further corroborated by the lack of significant difference ($p > .05$) in both the humans' and the robot's ER in the neutral condition compared to either of the other conditions (cf. Figures~\ref{fig:er_human_group} and~\ref{fig:er_robot_group}). The current study did not directly compare ER between humans and the robot under each condition. Because robots do not respond as accurately as humans, a lower accuracy is to be expected for robots~\cite{wang2020}. However, it is still important to find that the relevant values between incongruent and congruent conditions between humans and the robot are closely related. Although the robot shows significantly larger SRC effects than the humans, it is reasonable for responses from the robot to have more {\color{black}variability} than those of the humans. Though very low, the iCub's ego noise still makes audio localisation more challenging than for a human who could adjust to the visuals of the avatars in the pretrials. In contrast, the iCub could rely solely on its pretrained model. \textcolor{black}{Besides, although the participants respond to the stimuli by pressing the corresponding keys on the keyboard, while the iCub robot responds by shifting its eyes, the SRC effects still significantly show during the iCub experiment.} The robot provides a fixation density map, representing the most likely region a human would tend to fix his/her attention in a crossmodal audio-visual scenario. By providing different degrees of attention to each modality, guaranteeing that all of them would be considered for the determination of the fixation density map, {\color{black}the neurorobotic model is capable of generating the human-like crossmodal attention.} The possibility of making a humanoid robot mimic human attention behaviour is an essential step towards creating robots that can understand human intentions, predict their behaviours, and behave naturally and smoothly in human-robot interactions.

\section{Future Work}
\label{sec:futurework}

The current work could give way to studies from multiple areas and perspectives. For instance, during the social attention task, eye-tracking techniques could be used to collect human eye movement responses, e.g., pupil dilation, visual fixation, and microsaccades. This allows for a more comprehensive analysis of human attention under the different conditions of audio-visual congruency. Fine-tuning audio-visual saliency models on the collected task-specific data could lead to performance on par with humans.

To make the experimental design more diverse and realistic, future studies could utilise other social cues from the avatar's face and body. Besides, the experimental design could be enhanced by considering additional factors, such as the avatars' emotions and other identity features. This could be helpful for target speaker detection, emotion recognition, and sound localisation in future robotic studies. Considering that speaking activity is one key feature in determining which person to look at~\cite{xu2018}, it is crucial to consider when creating robots that mimic human attention behaviour. 

Our current work and findings can be applied to build social robots for interacting with children who have ASD or autistic traits. Previous research has shown that children with ASD avoid mutual gaze and other social interaction with humans, but not with humanoid robots~\cite{scassellati2012robots}. {\color{black}Children with ASD find humanoid robots with child-like appearances more approachable~\cite{shimaya2016advantages,raptopoulou2021human}}, which enables social robots to help them enhance their social functions.

Finally, the current experiment could be extended to a human-robot interaction scenario, such as replacing avatars with real humans or robots and evaluating responses from the participants and robots~\cite{andriella2020have}. There have been several human-robot interaction studies on how humans react to a robot's eye gaze~\cite{pfeifer2012operational,admoni2017social,willemse2018robot} or the mutual gaze effect on human decision-making~\cite{kompatsiari2021s,belkaid2021mutual}. Based on our study, what can be extended, but can also be challenging, is to make robots learn multiperson eye gaze and detect the active speaker in real-time during a collaborative task or social scenario with humans.

In conclusion, our interdisciplinary study provides new insights into how social cues trigger social attention in a complex multisensory scenario with realistic and dynamic social cues and stimuli. We also demonstrated that by predicting the fixation density map, the GASP model triggered the iCub robot to have a human-like response and similar socio-cognitive functions, resolving sensory conflicts within a high-level social context. By combining stimulus-driven information with internal targets and expectations, we hypothesise that these aspects of multisensory interaction should enable current computational models of robot perception to yield robust and flexible social behaviour during human-robot interaction.

\section{Acknowledgements}
We especially thank Dr. Cornelius Weber, Dr. Phillip Allgeuer, Dr. Zhong Yang, Dr. Guochun Yang, and Guangteng Meng for their improvement of the experimental design and the manuscript.

\section*{Author Contributions} 
D.F., X.L., and S.W. designed the experiment. D.F. and Z.C. collected the human data. F.A. conducted the computational modelling and robotic experiment. D.F. analysed the data. M.K. developed the framework for generating the experimental stimuli. M.K. and E.S. contributed to the experimental setup and stimuli generation. D.F., H.C., F.A., and M.K. wrote the manuscript. All authors contributed to improve the manuscript.

\section*{Compliance with Ethical Standards}
\textbf{Funding}: This work was supported by the National Natural Science Foundation of China (NSFC, No. 62061136001), the German Research Foundation (DFG) under project Transregio Crossmodal Learning (TRR 169). D.F. is funded by the Office of China Postdoctoral Council. 
\newline
\newline
\textbf{Conflict of Interest}: The authors declare that they have no conflict of interest.
\newline
\newline
\textbf{Ethical Standard}: All procedures performed in studies involving participants were following the ethical standards of the institutional and national research committee and with the 1964 Helsinki declaration and its later amendments or comparable ethical standards. 
\newline
\newline
\textbf{Informed Consent}: Informed consent was obtained from all participants in the study.

\section*{Data Availability Statement}
The datasets generated and analysed during the current study are available in the Open Science Framework repository, \url{https://osf.io/fbncu/}.
\bibliographystyle{spmpsci}      %
\bibliography{reference.bib}

\begin{thebibliography}{10}
\providecommand{\url}[1]{{#1}}
\providecommand{\urlprefix}{URL }
\expandafter\ifx\csname urlstyle\endcsname\relax
  \providecommand{\doi}[1]{DOI~\discretionary{}{}{}#1}\else
  \providecommand{\doi}{DOI~\discretionary{}{}{}\begingroup
  \urlstyle{rm}\Url}\fi

\bibitem{abawi2023wrapyfi}
Abawi, F., Allgeuer, P., Fu, D., Wermter, S.: {Wrapyfi: A Wrapper for
  Message-Oriented and Robotics Middleware} (2023).
\newblock \doi{10.48550/ARXIV.2302.09648}

\bibitem{abawi2021}
Abawi, F., Weber, T., Wermter, S.: {GASP}: Gated attention for saliency
  prediction.
\newblock In: Proceedings of the International Joint Conference on Artificial
  Intelligence ({IJCAI}), pp. 584--591. IJCAI Organization (2021).
\newblock \doi{10.24963/ijcai.2021/81}

\bibitem{admoni2017social}
Admoni, H., Scassellati, B.: Social eye gaze in human-robot interaction: a
  review.
\newblock Journal of Human-Robot Interaction \textbf{6}(1), 25--63 (2017).
\newblock \doi{10.5898/JHRI.6.1.Admoni}

\bibitem{akiyama2007unilateral}
Akiyama, T., Kato, M., Muramatsu, T., Umeda, S., Saito, F., Kashima, H.:
  Unilateral amygdala lesions hamper attentional orienting triggered by gaze
  direction.
\newblock Cerebral Cortex \textbf{17}(11), 2593--2600 (2007).
\newblock \doi{10.1093/cercor/bhl166}

\bibitem{ambrosecchia2015spatial}
Ambrosecchia, M., Marino, B.F., Gawryszewski, L.G., Riggio, L.: Spatial
  stimulus-response compatibility and affordance effects are not ruled by the
  same mechanisms.
\newblock Frontiers in human neuroscience \textbf{9}, 283 (2015).
\newblock \doi{10.3389/fnhum.2015.00283}

\bibitem{andriella2020have}
Andriella, A., Siqueira, H., Fu, D., Magg, S., Barros, P., Wermter, S., Torras,
  C., Alenya, G.: Do {I} have a personality? {E}ndowing care robots with
  context-dependent personality traits.
\newblock International Journal of Social Robotics pp. 1--22 (2020).
\newblock \doi{10.1007/s12369-020-00690-5}

\bibitem{arevalo2020}
Arevalo, J., Solorio, T., y~G{\'o}mez, M.M., Gonz{\'a}lez, F.A.: Gated
  multimodal networks.
\newblock Neural Computing and Applications \textbf{32}, 10209--10228 (2020).
\newblock \doi{10.1007/s00521-019-04559-1}

\bibitem{baron1997mindblindness}
Baron-Cohen, S.: Mindblindness: {A}n essay on autism and theory of mind.
\newblock MIT press (1997).
\newblock \doi{10.7551/mitpress/4635.001.0001}

\bibitem{battich2020coordinating}
Battich, L., Fairhurst, M., Deroy, O.: Coordinating attention requires
  coordinated senses.
\newblock Psychonomic Bulletin \& Review pp. 1--13 (2020).
\newblock \doi{10.3758/s13423-020-01766-z}

\bibitem{belkaid2021mutual}
Belkaid, M., Kompatsiari, K., De~Tommaso, D., Zablith, I., Wykowska, A.: Mutual
  gaze with a robot affects human neural activity and delays decision-making
  processes.
\newblock Science Robotics \textbf{6}(58), eabc5044 (2021).
\newblock \doi{10.1126/scirobotics.abc5044}

\bibitem{birmingham2009human}
Birmingham, E., Kingstone, A.: Human social attention: A new look at past,
  present, and future investigations.
\newblock Annals of the New York Academy of Sciences \textbf{1156}(1), 118--140
  (2009).
\newblock \doi{10.1111/j.1749-6632.2009.04468.x}

\bibitem{brooks2005development}
Brooks, R., Meltzoff, A.N.: The development of gaze following and its relation
  to language.
\newblock Developmental Science \textbf{8}(6), 535--543 (2005).
\newblock \doi{10.1111/j.1467-7687.2005.00445.x}

\bibitem{bylinskii2019eval}
{Bylinskii}, Z., {Judd}, T., {Oliva}, A., {Torralba}, A., {Durand}, F.: What do
  different evaluation metrics tell us about saliency models?
\newblock IEEE Transactions on Pattern Analysis and Machine Intelligence
  \textbf{41}(3), 740--757 (2019).
\newblock \doi{10.1109/TPAMI.2018.2815601}

\bibitem{cohen1990control}
Cohen, J.D., Dunbar, K., McClelland, J.L.: On the control of automatic
  processes: a parallel distributed processing account of the stroop effect.
\newblock Psychological Review \textbf{97}(3), 332 (1990).
\newblock \doi{10.1037/0033-295x.97.3.332}

\bibitem{cornia2018}
Cornia, M., Baraldi, L., Serra, G., Cucchiara, R.: Predicting human eye
  fixations via an {LSTM}-based saliency attentive model.
\newblock IEEE Transactions on Image Processing \textbf{27}(10), 5142--5154
  (2018).
\newblock \doi{10.1109/TIP.2018.2851672}

\bibitem{dalmaso2021face}
Dalmaso, M., Zhang, X., Galfano, G., Castelli, L.: Face masks do not alter gaze
  cueing of attention: Evidence from the covid-19 pandemic.
\newblock i-Perception \textbf{12}(6), 20416695211058480 (2021).
\newblock \doi{10.1177/20416695211058480}

\bibitem{doruk2018cross}
Doruk, D., Chanes, L., Malavera, A., Merabet, L.B., Valero-Cabr{\'e}, A.,
  Fregni, F.: Cross-modal cueing effects of visuospatial attention on conscious
  somatosensory perception.
\newblock Heliyon \textbf{4}(4), e00595 (2018).
\newblock \doi{10.1016/j.heliyon.2018.e00595}

\bibitem{eriksen1974effects}
Eriksen, B.A., Eriksen, C.W.: Effects of noise letters upon the identification
  of a target letter in a nonsearch task.
\newblock Perception \& Psychophysics \textbf{16}(1), 143--149 (1974).
\newblock \doi{10.3758/BF03203267}

\bibitem{farroni2004gaze}
Farroni, T., Massaccesi, S., Pividori, D., Johnson, M.H.: Gaze following in
  newborns.
\newblock Infancy \textbf{5}(1), 39--60 (2004).
\newblock \doi{10.1207/s15327078in0501_2}

\bibitem{friesen1998eyes}
Friesen, C.K., Kingstone, A.: The eyes have it! {R}eflexive orienting is
  triggered by nonpredictive gaze.
\newblock Psychonomic Bulletin \& Review \textbf{5}(3), 490--495 (1998).
\newblock \doi{10.3758/BF03208827}

\bibitem{friesen2004attentional}
Friesen, C.K., Ristic, J., Kingstone, A.: Attentional effects of
  counterpredictive gaze and arrow cues.
\newblock Journal of Experimental Psychology: Human Perception and Performance
  \textbf{30}(2), 319 (2004).
\newblock \doi{10.1037/0096-1523.30.2.319}

\bibitem{frischen2007gaze}
Frischen, A., Bayliss, A.P., Tipper, S.P.: Gaze cueing of attention: visual
  attention, social cognition, and individual differences.
\newblock Psychological Bulletin \textbf{133}(4), 694 (2007).
\newblock \doi{10.1037/0033-2909.133.4.694}

\bibitem{fu2018assessing}
Fu, D., Barros, P., Parisi, G.I., Wu, H., Magg, S., Liu, X., Wermter, S.:
  Assessing the contribution of semantic congruency to multisensory integration
  and conflict resolution.
\newblock In: IROS 2018 Workshop on Crossmodal Learning for Intelligent
  Robotics. IEEE (2018)

\bibitem{fu2020can}
Fu, D., Weber, C., Yang, G., Kerzel, M., Nan, W., Barros, P., Wu, H., Liu, X.,
  Wermter, S.: What can computational models learn from human selective
  attention? {A} review from an audiovisual unimodal and crossmodal
  perspective.
\newblock Frontiers in Integrative Neuroscience \textbf{14}, 10 (2020).
\newblock \doi{10.3389/fnint.2020.00010}

\bibitem{gao2019}
Gao, R., Grauman, K.: 2.5{D} visual sound.
\newblock In: Proceedings of the {IEEE/CVF} Conference on Computer Vision and
  Pattern Recognition ({CVPR}), pp. 324--333. IEEE (2019).
\newblock \doi{10.1109/CVPR.2019.00041}

\bibitem{gori2021masking}
Gori, M., Schiatti, L., Amadeo, M.B.: Masking emotions: face masks impair how
  we read emotions.
\newblock Frontiers in Psychology \textbf{12}, 1541 (2021).
\newblock \doi{10.3389/fpsyg.2021.669432}

\bibitem{guo2019abnormal}
Guo, J., Luo, X., Wang, E., Li, B., Chang, Q., Sun, L., Song, Y.: Abnormal
  alpha modulation in response to human eye gaze predicts inattention severity
  in children with {ADHD}.
\newblock Developmental Cognitive Neuroscience \textbf{38}, 100671 (2019).
\newblock \doi{10.1016/j.dcn.2019.100671}

\bibitem{hara2018can}
Hara, K., Kataoka, H., Satoh, Y.: Can spatiotemporal {3D CNNs} retrace the
  history of {2D CNNs} and {Imagenet}?
\newblock In: Proceedings of the {IEEE} conference on Computer Vision and
  Pattern Recognition ({CVPR}), pp. 6546--6555. IEEE (2018).
\newblock \doi{10.1109/CVPR.2018.00685}

\bibitem{he2015}
He, K., Zhang, X., Ren, S., Sun, J.: Delving deep into rectifiers: {S}urpassing
  human-level performance on {I}mage{N}et classification.
\newblock In: Proceedings of the IEEE International Conference on Computer
  Vision (ICCV), pp. 1026--1034. IEEE, USA (2015).
\newblock \doi{10.1109/ICCV.2015.123}

\bibitem{he2016identity}
He, K., Zhang, X., Ren, S., Sun, J.: Identity mappings in deep residual
  networks.
\newblock In: European conference on computer vision, pp. 630--645. Springer
  (2016).
\newblock \doi{10.1007/978-3-319-46493-0_38}

\bibitem{zeromq2013hintjens}
Hintjens, P.: ZeroMQ: messaging for many applications.
\newblock "O'Reilly Media, Inc." (2013).
\newblock \urlprefix\url{https://zeromq.org/}

\bibitem{jain2020vinet}
Jain, S., Yarlagadda, P., Jyoti, S., Karthik, S., Subramanian, R., Gandhi, V.:
  {ViNet}: Pushing the limits of visual modality for audio-visual saliency
  prediction.
\newblock In: Proceedings of the IEEE/RSJ International Conference on
  Intelligent Robots and Systems (IROS), pp. 3520--3527. IEEE (2020).
\newblock \doi{10.1109/IROS51168.2021.9635989}

\bibitem{jessen2014unconscious}
Jessen, S., Grossmann, T.: Unconscious discrimination of social cues from eye
  whites in infants.
\newblock Proceedings of the National Academy of Sciences \textbf{111}(45),
  16208--16213 (2014).
\newblock \doi{10.1073/pnas.1411333111}

\bibitem{johnson1998whose}
Johnson, S., Slaughter, V., Carey, S.: Whose gaze will infants follow? the
  elicitation of gaze-following in 12-month-olds.
\newblock Developmental Science \textbf{1}(2), 233--238 (1998).
\newblock \doi{10.1111/1467-7687.00036}

\bibitem{kerzel2020}
Kerzel, M., Wermter, S.: Towards a data generation framework for affective
  shared perception and social cue learning using virtual avatars.
\newblock In: Workshop on Affective Shared Perception, ICDL 2020, {IEEE}
  International Conference on Development and Learning (2020)

\bibitem{kompatsiari2021s}
Kompatsiari, K., Ciardo, F., Tikhanoff, V., Metta, G., Wykowska, A.: It’s in
  the eyes: The engaging role of eye contact in {HRI}.
\newblock International Journal of Social Robotics \textbf{13}(3), 525--535
  (2021).
\newblock \doi{10.1007/s12369-019-00565-4}

\bibitem{kornblum1995stimulus}
Kornblum, S., Lee, J.W.: Stimulus-response compatibility with relevant and
  irrelevant stimulus dimensions that do and do not overlap with the response.
\newblock Journal of Experimental Psychology: Human Perception and Performance
  \textbf{21}(4), 855 (1995)

\bibitem{langton2000eyes}
Langton, S.R., Watt, R.J., Bruce, V.: Do the eyes have it? {C}ues to the
  direction of social attention.
\newblock Trends in Cognitive Sciences \textbf{4}(2), 50--59 (2000).
\newblock \doi{10.1016/s1364-6613(99)01436-9}

\bibitem{laube2011cortical}
Laube, I., Kamphuis, S., Dicke, P.W., Thier, P.: Cortical processing of
  head-and eye-gaze cues guiding joint social attention.
\newblock Neuroimage \textbf{54}(2), 1643--1653 (2011).
\newblock \doi{10.1016/j.neuroimage.2010.08.074}

\bibitem{liu2018neurodevelopment}
Liu, X., Liu, T., Shangguan, F., S{\o}rensen, T.A., Liu, Q., Shi, J.:
  Neurodevelopment of conflict adaptation: Evidence from event-related
  potentials.
\newblock Developmental Psychology \textbf{54}(7), 1347 (2018).
\newblock \doi{10.1037/dev0000524}

\bibitem{macleod1991half}
MacLeod, C.M.: Half a century of research on the stroop effect: an integrative
  review.
\newblock Psychological bulletin \textbf{109}(2), 163 (1991)

\bibitem{maddox2014directing}
Maddox, R.K., Pospisil, D.A., Stecker, G.C., Lee, A.K.: Directing eye gaze
  enhances auditory spatial cue discrimination.
\newblock Current Biology \textbf{24}(7), 748--752 (2014).
\newblock \doi{10.1016/j.cub.2014.02.021}

\bibitem{mcneely2003neurophysiological}
McNeely, H.E., West, R., Christensen, B.K., Alain, C.: Neurophysiological
  evidence for disturbances of conflict processing in patients with
  schizophrenia.
\newblock Journal of Abnormal Psychology \textbf{112}(4), 679 (2003).
\newblock \doi{10.1037/0021-843X.112.4.679}

\bibitem{yarp2006metta}
Metta, G., Fitzpatrick, P., Natale, L.: Yarp: yet another robot platform.
\newblock International Journal of Advanced Robotic Systems \textbf{3}(1), 8
  (2006).
\newblock \doi{10.5772/5761}

\bibitem{mundy2007attention}
Mundy, P., Newell, L.: Attention, joint attention, and social cognition.
\newblock Current Directions in Psychological Science \textbf{16}(5), 269--274
  (2007).
\newblock \doi{10.1111/j.1467-8721.2007.00518.x}

\bibitem{newport2009short}
Newport, R., Howarth, S.: Social gaze cueing to auditory locations.
\newblock Quarterly Journal of Experimental Psychology \textbf{62}(4), 625--634
  (2009).
\newblock \doi{10.1080/17470210802486027}

\bibitem{nocentini2019survey}
Nocentini, O., Fiorini, L., Acerbi, G., Sorrentino, A., Mancioppi, G., Cavallo,
  F.: A survey of behavioral models for social robots.
\newblock Robotics \textbf{8}(3), 54 (2019).
\newblock \doi{10.3390/robotics8030054}

\bibitem{nuku2008joint}
Nuku, P., Bekkering, H.: Joint attention: Inferring what others perceive (and
  don’t perceive).
\newblock Consciousness and Cognition \textbf{17}(1), 339--349 (2008).
\newblock \doi{10.1016/j.concog.2007.06.014}

\bibitem{nuku2010one}
Nuku, P., Bekkering, H.: When one sees what the other hears: {C}rossmodal
  attentional modulation for gazed and non-gazed upon auditory targets.
\newblock Consciousness and Cognition \textbf{19}(1), 135--143 (2010).
\newblock \doi{10.1016/j.concog.2009.07.012}

\bibitem{nummenmaa2009}
Nummenmaa, L., Calder, A.J.: Neural mechanisms of social attention.
\newblock Trends in Cognitive Sciences \textbf{13}(3), 135--143 (2009).
\newblock \doi{10.1016/j.tics.2008.12.006}

\bibitem{parisi2018neurorobotic}
Parisi, G.I., Barros, P., Fu, D., Magg, S., Wu, H., Liu, X., Wermter, S.: A
  neurorobotic experiment for crossmodal conflict resolution in complex
  environments.
\newblock In: Proceedings of the IEEE/RSJ International Conference on
  Intelligent Robots and Systems (IROS), pp. 2330--2335. IEEE (2018).
\newblock \doi{10.1109/IROS.2018.8594036}

\bibitem{pfeifer2012operational}
Pfeifer-Lessmann, N., Pfeifer, T., Wachsmuth, I.: An operational model of joint
  attention-timing of gaze patterns in interactions between humans and a
  virtual human.
\newblock In: Proceedings of the Annual Meeting of the Cognitive Science
  Society, vol.~34 (2012)

\bibitem{posner1984components}
Posner, M., Cohen, Y.: Components of visual orienting.
\newblock In: Attention and performance X: Control of language processes, pp.
  531--556. Psychology Press, London, United Kingdom (1984)

\bibitem{posner1980attention}
Posner, M.I., Snyder, C.R., Davidson, B.J.: Attention and the detection of
  signals.
\newblock Journal of Experimental Psychology: General \textbf{109}(2), 160
  (1980)

\bibitem{proctor2006stimulus}
Proctor, R.W., Vu, K.P.L.: Stimulus-response compatibility principles: Data,
  theory, and application.
\newblock CRC press (2006)

\bibitem{rachavarapu2021localize}
Rachavarapu, K.K., Sundaresha, V., Aakanksha, Rajagopalan, A.: Localize to
  binauralize: Audio spatialization from visual sound source localization.
\newblock In: Proceedings of the IEEE/CVF International Conference on Computer
  Vision, pp. 1930--1939. IEEE (2021).
\newblock \doi{10.1109/ICCV48922.2021.00194}

\bibitem{raptopoulou2021human}
Raptopoulou, A., Komnidis, A., Bamidis, P.D., Astaras, A.: Human--robot
  interaction for social skill development in children with asd: A literature
  review.
\newblock Healthcare Technology Letters \textbf{8}(4), 90--96 (2021)

\bibitem{ristic2007attentional}
Ristic, J., Wright, A., Kingstone, A.: Attentional control and reflexive
  orienting to gaze and arrow cues.
\newblock Psychonomic Bulletin \& Review \textbf{14}(5), 964--969 (2007).
\newblock \doi{10.3758/bf03194129}

\bibitem{scassellati2012robots}
Scassellati, B., Admoni, H., Matari{\'c}, M.: Robots for use in autism
  research.
\newblock Annual Review of Biomedical Engineering \textbf{14}, 275--294 (2012).
\newblock \doi{10.1146/annurev-bioeng-071811-150036}

\bibitem{schuller2004perception}
Schuller, A.M., Rossion, B.: Perception of static eye gaze direction
  facilitates subsequent early visual processing.
\newblock Clinical Neurophysiology \textbf{115}(5), 1161--1168 (2004)

\bibitem{senju2009atypical}
Senju, A., Johnson, M.H.: Atypical eye contact in autism: models, mechanisms
  and development.
\newblock Neuroscience \& Biobehavioral Reviews \textbf{33}(8), 1204--1214
  (2009).
\newblock \doi{10.1016/j.neubiorev.2009.06.001}

\bibitem{shepherd2010following}
Shepherd, S.V.: Following gaze: {G}aze-following behavior as a window into
  social cognition.
\newblock Frontiers in Integrative Neuroscience \textbf{4}, 5 (2010).
\newblock \doi{10.3389/fnint.2010.00005}

\bibitem{shimaya2016advantages}
Shimaya, J., Yoshikawa, Y., Matsumoto, Y., Kumazaki, H., Ishiguro, H., Mimura,
  M., Miyao, M.: Advantages of indirect conversation via a desktop humanoid
  robot: Case study on daily life guidance for adolescents with autism spectrum
  disorders.
\newblock In: 2016 25th IEEE international symposium on robot and human
  interactive communication (RO-MAN), pp. 831--836. IEEE (2016)

\bibitem{simon1967auditory}
Simon, J.R., Rudell, A.P.: Auditory sr compatibility: the effect of an
  irrelevant cue on information processing.
\newblock Journal of applied psychology \textbf{51}(3), 300 (1967)

\bibitem{soto2005spatial}
Soto-Faraco, S., Sinnett, S., Alsius, A., Kingstone, A.: Spatial orienting of
  tactile attention induced by social cues.
\newblock Psychonomic Bulletin \& Review \textbf{12}(6), 1024--1031 (2005).
\newblock \doi{10.3758/BF03206438}

\bibitem{sperdin2018}
Sperdin, H.F., Coito, A., Kojovic, N., Rihs, T.A., Jan, R.K., Franchini, M.,
  Plomp, G., Vulliemoz, S., Eliez, S., Michel, C.M., Schaer, M.: Early
  alterations of social brain networks in young children with autism.
\newblock eLife \textbf{7}, 1--23 (2018).
\newblock \doi{10.7554/eLife.31670}

\bibitem{srinivasan2016effects}
Srinivasan, S.M., Eigsti, I.M., Neelly, L., Bhat, A.N.: The effects of embodied
  rhythm and robotic interventions on the spontaneous and responsive social
  attention patterns of children with autism spectrum disorder (asd): A pilot
  randomized controlled trial.
\newblock Research in autism spectrum disorders \textbf{27}, 54--72 (2016)

\bibitem{stajduhar2022face}
Stajduhar, A., Ganel, T., Avidan, G., Rosenbaum, R.S., Freud, E.: Face masks
  disrupt holistic processing and face perception in school-age children.
\newblock Cognitive research: principles and implications \textbf{7}(1), 1--10
  (2022).
\newblock \doi{10.1186/s41235-022-00360-2}

\bibitem{stroop1935studies}
Stroop, J.R.: Studies of interference in serial verbal reactions.
\newblock Journal of experimental psychology \textbf{18}(6), 643 (1935)

\bibitem{tavakoli2020}
Tavakoli, H.R., Borji, A., Kannala, J., Rahtu, E.: Deep audio-visual saliency:
  {B}aseline model and data.
\newblock In: ACM Symposium on Eye Tracking Research and Applications, ETRA '20
  Short Papers, pp. 1--5. Association for Computing Machinery, New York, NY,
  USA (2020).
\newblock \doi{10.1145/3379156.3391337}

\bibitem{tsiami2020stavis}
Tsiami, A., Koutras, P., Maragos, P.: {STAViS}: Spatio-temporal audiovisual
  saliency network.
\newblock In: Proceedings of the {IEEE/CVF} Conference on Computer Vision and
  Pattern Recognition (CVPR), pp. 4766--4776. IEEE (2020).
\newblock \doi{10.1109/CVPR42600.2020.00482}

\bibitem{wang2020}
Wang, J., Wang, J., Qian, K., Xie, X., Kuang, J.: Binaural sound localization
  based on deep neural network and affinity propagation clustering in
  mismatched {HRTF} condition.
\newblock {EURASIP} Journal on Audio, Speech, and Music Processing
  \textbf{2020}(4), 1--16 (2020).
\newblock \doi{10.1186/s13636-020-0171-y}

\bibitem{wightman1997monaural}
Wightman, F.L., Kistler, D.J.: Monaural sound localization revisited.
\newblock The Journal of the Acoustical Society of America \textbf{101}(2),
  1050--1063 (1997)

\bibitem{willemse2018robot}
Willemse, C., Marchesi, S., Wykowska, A.: Robot faces that follow gaze
  facilitate attentional engagement and increase their likeability.
\newblock Frontiers in Psychology \textbf{9}, 70 (2018).
\newblock \doi{10.3389/fpsyg.2018.00070}

\bibitem{wu2021binaural}
Wu, X., Wu, Z., Ju, L., Wang, S.: Binaural Audio-Visual Localization, vol.
  35(4).
\newblock AAAI (2021)

\bibitem{xu2018}
Xu, M., Liu, Y., Hu, R., He, F.: Find who to look at: {T}urning from action to
  saliency.
\newblock {IEEE} Transactions on Image Processing \textbf{27}(9), 4529--4544
  (2018).
\newblock \doi{10.1109/TIP.2018.2837106}

\bibitem{yeung2013lip}
Yeung, H.H., Werker, J.F.: Lip movements affect infants’ audiovisual speech
  perception.
\newblock Psychological Science \textbf{24}(5), 603--612 (2013)

\end{thebibliography}
\appendix
\section*{Supplementary Materials}
The example of experimental stimuli and videos for both human data and the iCub robot data collection can be viewed at this link:\\ \url{https://www.youtube.com/watch?v=bjiYEs1x-7E}.
\label{app:gsq}
\end{document}